\documentclass[10pt,twocolumn,letterpaper]{article}
\usepackage[utf8]{inputenc}
\DeclareUnicodeCharacter{3002}{.}

\usepackage{iccv}
\usepackage{times}
\usepackage{epsfig}
\usepackage{graphicx}
\usepackage{amsmath}
\usepackage{amssymb}
\usepackage{array}
\usepackage{rotating}
\usepackage{makecell}
\usepackage{graphicx}
\usepackage{booktabs}
\usepackage{siunitx}
\usepackage{multirow}
\usepackage{xcolor}
\usepackage{verbatim}

\usepackage{colortbl}
\usepackage{array}
\usepackage{caption}
\usepackage{adjustbox}

\usepackage[breaklinks=true,bookmarks=false]{hyperref}

\iccvfinalcopy % *** Uncomment this line for the final submission

\def\iccvPaperID{1988} % *** Enter the ICCV Paper ID here

\definecolor{color3}{rgb}{0.95,0.95,0.95}

\definecolor{first}{HTML}{F7E1ED} % 浅蓝色F7E1ED
\definecolor{second}{HTML}{E6EFFA} % 浅蓝色

\newcommand{\lwb}[1]{{\color{blue}{#1}}}
\newcommand{\jingjing}[1]{{\color{red}{#1}}}

\definecolor{lightgray}{rgb}{0.95, 0.95, 0.95}
\definecolor{mediumgray}{rgb}{0.85, 0.85, 0.85}
\definecolor{tableheadcolor}{rgb}{0.7, 0.7, 0.7}
\definecolor{modelcolor}{rgb}{0.2, 0.4, 0.6}

\ificcvfinal\fi

\begin{document}

%%%%%%%%% TITLE
\title{Turbo2K: Towards Ultra-Efficient and High-Quality 2K Video Synthesis}

\author{
Jingjing Ren$^{1\dagger}$,\quad Wenbo Li$^{2\dagger}$, \quad Zhongdao Wang$^{2}$, \quad Haoze Sun$^{2}$, \quad Bangzhen Liu$^{3}$, \quad
Haoyu Chen$^{1}$, \\ \quad Jiaqi Xu$^{2}$, \quad Aoxue Li$^{2}$, \quad Shifeng Zhang$^{2}$ , \quad Bin Shao$^{2}$, \quad Yong Guo$^{4}$, \quad Lei Zhu$^{1,5*}$ \\ \vspace{-0.5mm}
\footnotesize $^{1}$The Hong Kong University of Science and Technology (Guangzhou)\quad
\footnotesize $^{2}$Huawei Noah’s Ark Lab
\vspace{-0.5mm}  
\footnotesize $^{3}$South China University of Technology \quad \\
\footnotesize $^{4}$Max Planck Institute for Informatics \quad
\footnotesize $^{5}$The Hong Kong University of Science and Technology\\
{\tt\small Project page: \url{https://jingjingrenabc.github.io/turbo2k/}}\\
\footnotesize $^\dagger$Equal contribution \quad $^*$Corresponding author
}

\twocolumn[{%
\renewcommand\twocolumn[1][]{#1}%
\maketitle
\vspace{-7mm}

\includegraphics[width=1\linewidth]{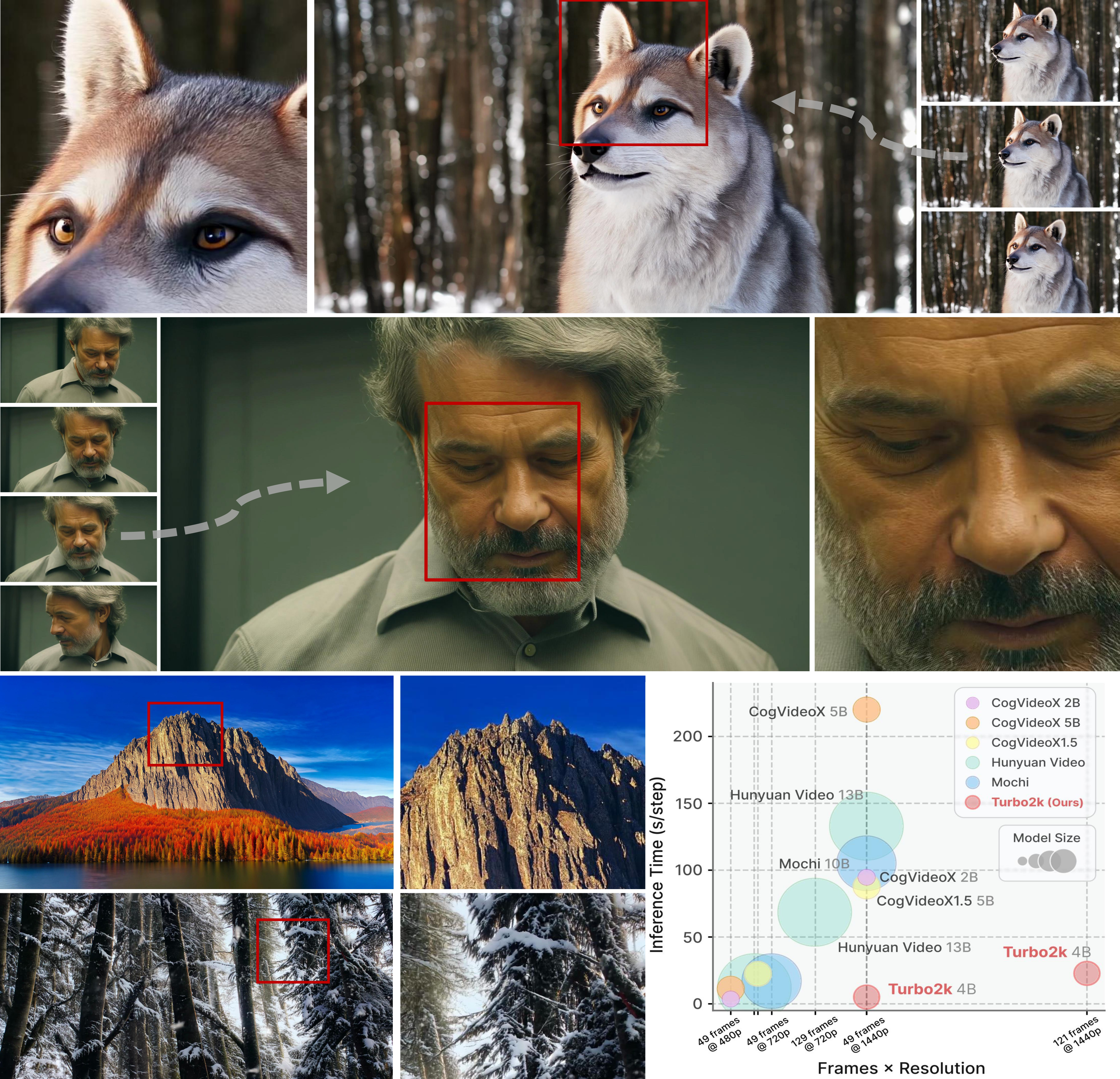}
\vspace{-7mm}
\captionof{figure}{Turbo2K generates high-quality, detail-rich, and aesthetically pleasing videos while achieving significant speed advantages over existing methods. Please refer to our supplementary file for more videos.}
\vspace{1em}
\label{fig:teaser}
}]

%%%%%%%%% ABSTRACT
\begin{abstract}
Demand for 2K video synthesis is rising with increasing consumer expectations for ultra-clear visuals.
While diffusion transformers (DiTs) have demonstrated remarkable capabilities in high-quality video generation, scaling them to 2K resolution remains computationally prohibitive due to quadratic growth in memory and processing costs.
In this work, we propose Turbo2K, an efficient and practical framework for generating detail-rich 2K videos while significantly improving training and inference efficiency. First, Turbo2K operates in a highly compressed latent space, reducing computational complexity and memory footprint, making high-resolution video synthesis feasible. However, the high compression ratio of the VAE and limited model size impose constraints on generative quality. To mitigate this, we introduce a  knowledge distillation strategy that enables a smaller student model to inherit the generative capacity of a larger, more powerful teacher model. 
Our analysis reveals that, despite differences in latent spaces and architectures, DiTs exhibit structural similarities in their internal representations, facilitating effective knowledge transfer.
Second, we design a hierarchical two-stage synthesis framework that first generates multi-level feature at lower resolutions before guiding high-resolution video generation. This approach ensures structural coherence and fine-grained detail refinement while eliminating redundant encoding-decoding overhead, further enhancing computational efficiency.
Turbo2K achieves state-of-the-art efficiency, generating 5-second, 24fps, 2K videos with significantly reduced computational cost. Compared to existing methods, Turbo2K is up to 20$\times$ faster for inference, making high-resolution video generation more scalable and practical for real-world applications.
\end{abstract}

%%%%%%%%% BODY TEXT
\section{Introduction}

Diffusion Transformers (DiTs)~\cite{peebles2023scalable} have demonstrated remarkable capabilities in generating visually compelling images~\cite{chen2025pixart, esser2024scaling, zhuo2024lumina} and videos~\cite{kong2024hunyuanvideo, hacohen2024ltx, hong2022cogvideo, yang2024cogvideox}, benefiting from their scalability in both parameters and data. 
However, as display technologies evolve and consumer expectations for ultra-high-definition content surge, there is an urgent need for generative models that can produce high-resolution videos—such as 2K content—even though doing so remains challenging due to the immense computational overhead involved in both training and inference.
\begin{figure}[t]
  \centering
   \includegraphics[width=1\linewidth]{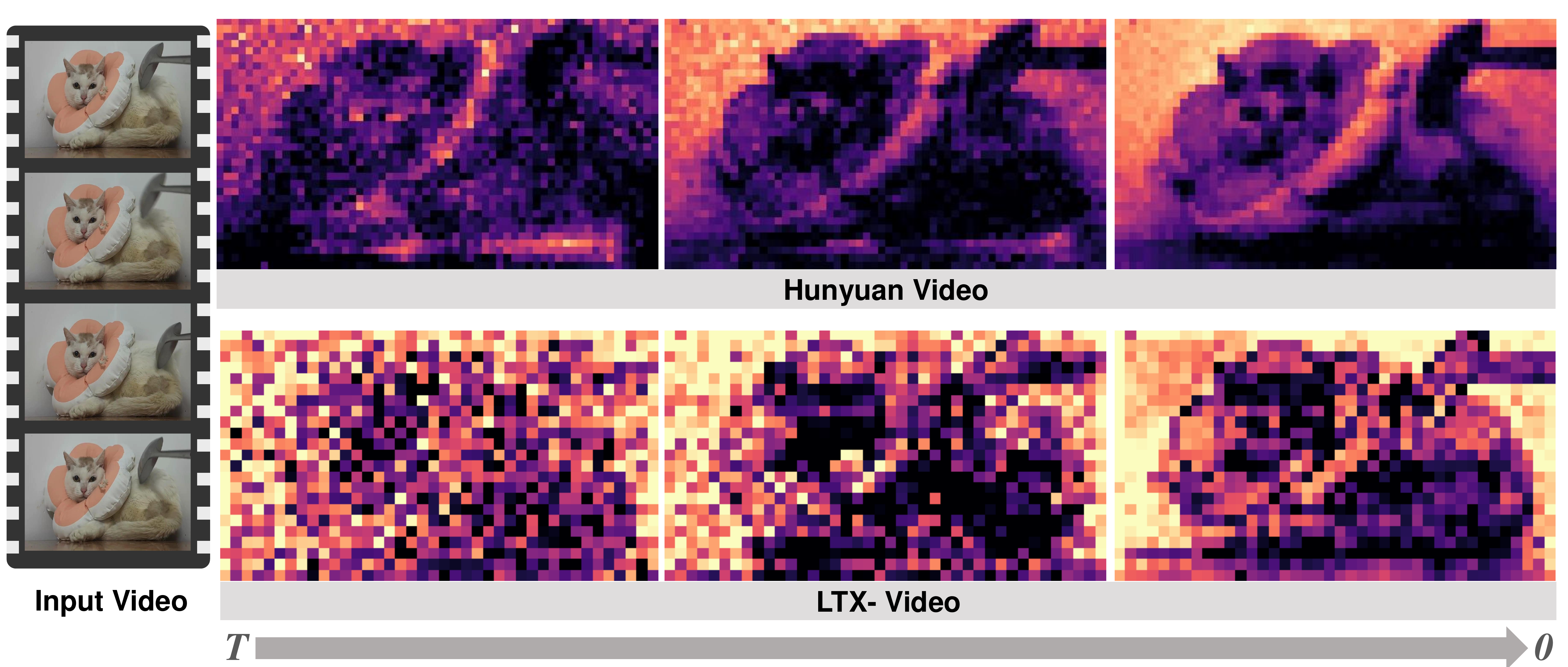}
   \caption{Visualization of internal feature structures across timesteps for different video diffusion models. The features exhibit similar underlying semantic structures.}
   \vspace{-0.15in}
   \label{fig:feature_vis}
\end{figure}

Within text-to-video (T2V) DiTs, the generation process operates in a compact latent space defined by a VAE~\cite{esser2021taming, kingma2013auto}, where transformer blocks~\cite{peebles2023scalable} seamlessly integrate textual and visual tokens. Yet, the quadratic scaling of the attention cost with token count imposes significant limitations on high-resolution generation. Techniques such as decoupled spatial-temporal attention~\cite{fan2025vchitect}, linear attention variants~\cite{xie2024sana, liu2024linfusion}, token compression~\cite{chen2025pixart}, localized attention~\cite{liu2024clear}, and quantization~\cite{zhang2411sageattention2} have been proposed to alleviate this issue, but these approaches often sacrifice video quality and frame consistency in complex scenarios. In contrast, we advocate for employing VAEs with high compression ratios that still enable high-quality reconstruction~\cite{agarwal2025cosmos, hacohen2024ltx, chen2024deep}, thereby enhancing training efficiency and reducing computational demands. 
\begin{figure*}[t]
  \centering
   \includegraphics[width=1\linewidth]{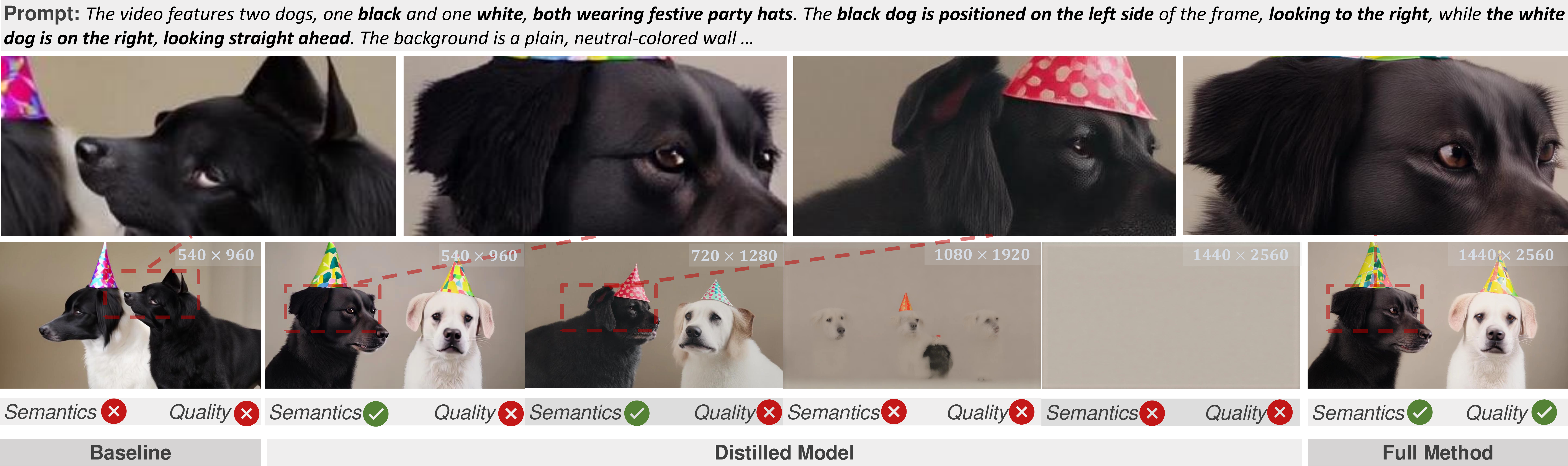}
   \caption{Generated results at different resolutions. The distilled base model performs well at 540p and 720p but degrades at higher resolutions, while our method maintains rich details and structural coherence at 2K resolution.}
   \label{fig:multires_compare}
   \vspace{-0.1in}
\end{figure*}

Mainstream video diffusion models~\cite{kong2024hunyuanvideo, hong2022cogvideo, lin2024open} typically utilize VAEs that compress videos with a $8 \times 8 \times 4$ factor spatial-temporally, achieving impressive generative performance at 540p or 720p. Conversely, LTX-Video~\cite{hacohen2024ltx} leverages a VAE with a substantially higher compression ratio, attaining an $32 \times 32 \times 8$ compression. By bypassing the additional tokenization step into DiTs, LTX-Video produces eight times fewer tokens than its counterparts, markedly reducing visual complexity and computational cost~\cite{pernias2023wurstchen}. Nevertheless, despite these advantages, LTX-Video's generative quality and stability are comparatively constrained when evaluated against other large-scale models~\cite{kong2024hunyuanvideo, genmo2024mochi, yang2024cogvideox} with significantly higher parameter counts.

To address the challenge of 2K video synthesis, we introduce Turbo2K, a framework that significantly enhances both generation quality and efficiency. Turbo2K can generate 5-second, 24fps, 2K videos while achieving notable speed improvements. Specifically, for 49-frame 2K video generation, Turbo2K runs approximately 20 times faster than CogVideoX~\cite{hong2022cogvideo} and HunyuanVideo~\cite{kong2024hunyuanvideo} as shown in Fig.~\ref{fig:teaser}. Moreover, Turbo2K is the only method capable of generating 121-frame 2K videos on a single A100 GPU, a feat that existing methods fail to achieve due to their high computational requirements.
By leveraging LTX-Video's VAE for efficient token compression, Turbo2K reduces computational costs. However, generating high-resolution videos requires fine-grained textures and consistent frame structures, which are challenging for smaller models like the 2B-parameter DiT. To resolve this, we employ knowledge distillation to transfer the generative capabilities of large models to smaller ones, allowing them to produce high-quality content.
As shown in Fig.~\ref{fig:feature_vis}, despite differences in latent spaces and architectures, video generative models share similar semantic structures and layouts, with DiT representations across timesteps displaying consistent patterns. 
This structural alignment would enables effective knowledge transfer through distillation, even when the teacher and student models, or their VAEs, differ. Additionally, the teacher's smooth data space simplifies the student's learning process, beneficial to semantic planning, while its lower compression rate preserves finer details, allowing the student to capture more intricate structural information. As illustrated in Fig.~\ref{fig:multires_compare}, the baseline model (LTX-Video) generates less coherent semantics and degraded details. In contrast, the distilled model produces significantly more plausible and visually refined outputs at the same resolution.

Nonetheless, even with enhanced generative capabilities, the distilled model remains sensitive to training resolution and struggles to scale to 2K resolution. As shown in Fig.~\ref{fig:multires_compare}, the 1080p output exhibits noticeable pattern inconsistencies, and the 2K result exhibits extensive regions of uniform color and diminished details.
Direct training at 2K resolution introduces significant challenges in both semantic planning and detail generation. To overcome these issues, we propose a two-stage framework: initially generating a low-resolution video from which semantic guidance is extracted to help high-resolution synthesis. Specifically, we leverage multi-level intermediate representations from DiTs to guide the subsequent high-resolution generation stage. This guidance-based approach, rather than direct latent upsampling, seamlessly connects the stages, eliminating the need for additional encoding/decoding steps and bridging the simulation gap associated with low-resolution inputs during training and inference. Consequently, our design substantially reduces the difficulty of high-resolution video synthesis, cutting computational costs and enabling the model to focus on producing visually appealing details. 

Our full method, as shown in Fig.~\ref{fig:multires_compare}, effectively synthesizes 2K videos with strong semantic coherence and visually pleasing fine-grained details. Our contributions can be summarized as follows: 
\begin{itemize}
\item \textbf{Turbo2K for 2K video generation.} We introduce Turbo2K, a robust framework that significantly elevates both the quality and efficiency of 2K video synthesis. It generates 5-second, 24fps, 2K videos at speeds that far exceed current methods while delivering exceptional visual quality.
\item \textbf{Heterogeneous model distillation.} We propose a novel distillation approach that effectively and efficiently transfers knowledge from large models operating in diverse VAE spaces to smaller models, thereby markedly enhancing their generative performance.
\item \textbf{Hierarchical two-stage framework.} We develop a hierarchical, two-stage framework that employs semantically derived low-resolution guidance to drive the high-resolution generation process, producing fine-grained and consistent details.
\end{itemize}

\section{Related Work}
\noindent{\textbf{Text-to-video diffusion models.}} 
Video diffusion models~\cite{hong2022cogvideo, yang2024cogvideox, kong2024hunyuanvideo, lin2024open, zheng2024open} have demonstrated the ability to generate high-quality video clips from text prompts. These models typically consist of two core components: a variational autoencoder (VAE) for spatial-temporal compression and a diffusion transformer (DiT)~\cite{peebles2023scalable} for sequential denoising.
Most video diffusion models employ a VAE with a compression ratio of $8\times 8 \times 4$ (spatial × spatial × temporal), meaning the video is temporally compressed by a factor of 4 and spatially by a factor of 8. Additionally, the DiT further processes latent representation using a $2 \times 2$ patch embedding, significantly lowering the number of tokens used for denoising.
A key distinction of LTX-Video~\cite{hacohen2024ltx} from other video generative models is its extremely high compression rate, which reduces the temporal dimension by a factor of 8 and the spatial dimensions by a factor of 32. This aggressive compression strategy enables significant computational efficiency, reducing the overall number of latent tokens while still maintaining reasonable generative quality.

\begin{figure*}[t]
  \centering
   \includegraphics[width=1\linewidth]{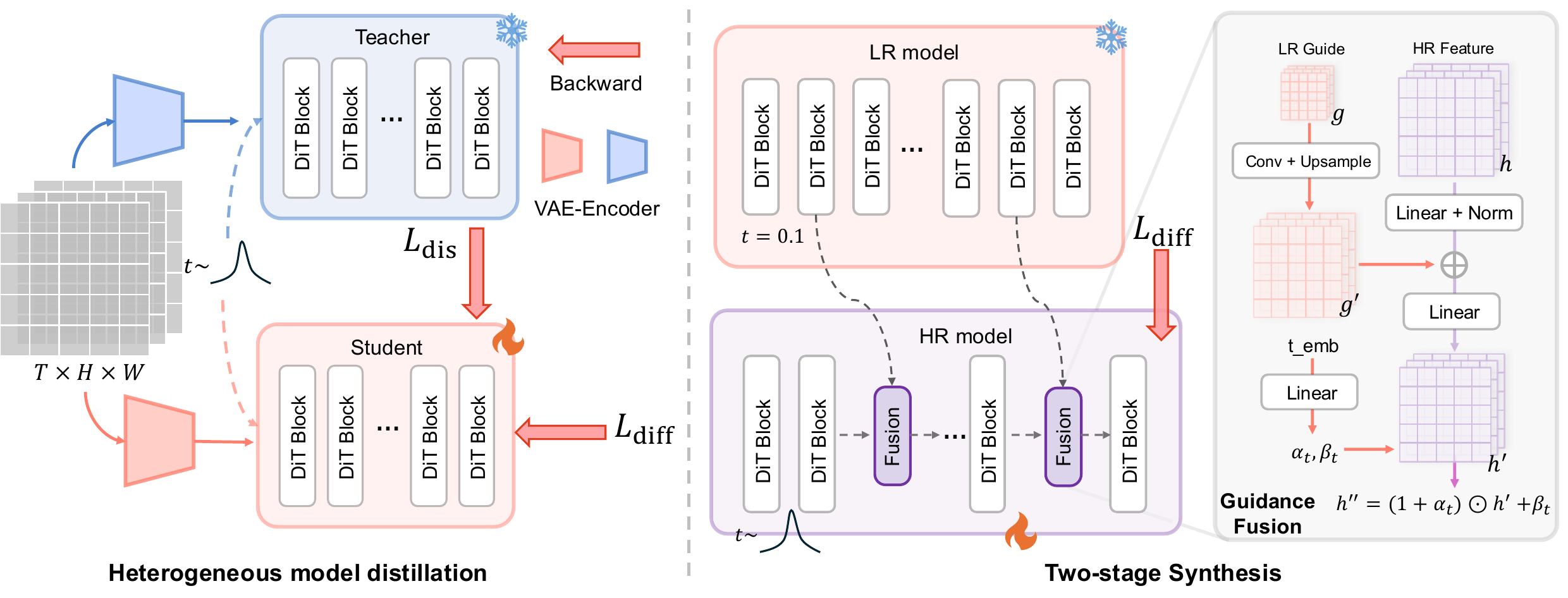}
   \caption{Turbo2K framework overview. \textbf{Left:} Heterogeneous model distillation aligns the student model’s internal representation with a larger teacher model to enhance semantic understanding and detail richness. \textbf{
   Right:} Two-stage synthesis first generates a low-resolution (LR) video, extracting semantic features to guide high-resolution (HR) generation.}
   \label{fig:method}
\end{figure*}

\noindent\textbf{High resolution image and video synthesis.} 
Generating high-resolution images and videos presents several challenges, including increased computational complexity, the need for high-quality training data, and scalability limitations. Existing approaches address these issues through tuning-free methods, fine-tuning strategies, and super-resolution techniques.
One line of work explores tuning-free approaches~\cite{he2023scalecrafter, du2024demofusion, liu2024hiprompt, zhang2023hidiffusion}, which enhance base diffusion models by modifying denoising strategies and architectural components, enabling higher-resolution generation without additional fine-tuning. Another approach adapts pre-trained generative models to higher resolutions via full or partial fine-tuning~\cite{cheng2024resadapter, ren2024ultrapixel, chen2025pixart, guo2025make}, improving output resolution while preserving model efficiency.
Super-resolution (SR)-based methods~\cite{he2024venhancer, zhou2024upscale, wang2025seedvr, wu2024seesr, wang2024exploiting, yu2024scaling} leverage the generative power of diffusion models to upscale low-resolution (LR) images and videos by treating LR inputs as conditions for generating high-resolution outputs. In the video domain, video SR methods typically process LR videos using overlapped patches or modify self-attention mechanisms~\cite{liu2021swin}, but they often suffer from high inference costs and limited generative flexibility due to their dependence on patch-wise constraints.
Our method differs from SR-based approaches by directly generating high-resolution videos with a two-stage synthesis framework, ensuring rich details and strong coherence while maintaining computational efficiency.

\section{Method}
Generating 2K videos poses significant challenges due to the need for both high-level semantic planning and fine-grained detail synthesis. Moreover, training such a model demands extensive computational resources.
To enable efficient training, we first enhance the model’s internal representation by aligning it with that of a larger teacher model, which operates at a lower compression rate. This alignment improves both semantic understanding and detail richness, as illustrated in the left part of Fig.~\ref{fig:method}.
Building on the enhanced distilled model, we introduce a two-stage synthesis pipeline, where the final 2K video generation is guided by semantic features extracted from the low-resolution (LR) generation process. This guidance mechanism ensures structural coherence and fine detail richness, as depicted in the right part of Fig.~\ref{fig:method}.
In the following, Sec.~\ref{sec:method_distill} introduces the proposed distillation framework, while Sec.~\ref{sec:method_twostage} presents the two-stage synthesis.

\subsection{Heterogeneous Model Distillation}
High-resolution video generation remains computationally intensive, primarily due to the quadratic scaling of DiT’s full attention mechanism with respect to the number of tokens. For instance, generating a 2K-resolution video comprising 49 frames using CogVideoX-5B requires approximately 100 seconds per denoising step on an NVIDIA A100 GPU, as shown in Fig.~\ref{fig:teaser}. To alleviate computational overhead, we adopt LTX-Video\cite{hacohen2024ltx}'s VAE design, which reduces the token count by 8$\times$~\cite{hong2022cogvideo, yang2024cogvideox, lin2024open}. However, despite the improved efficiency, its generative capacity remains significantly limited when compared to larger video diffusion models~\cite{kong2024hunyuanvideo, genmo2024mochi}.
A straightforward strategy to enhance generative performance is data-centric fine-tuning, where the model is further trained on a curated dataset. However, under moderate computational budgets, this approach yields limited improvements in output quality (see Table~\ref{tab:abla_distill} and Fig.~\ref{fig:distill_compare}). These results suggest that simply increasing the amount of training data is insufficient to bridge the performance gap between LTX and larger-scale video diffusion models, highlighting the necessity of more effective knowledge transfer mechanisms.

To address this challenge, we introduce a distillation framework wherein a lightweight, LTX-based student model is trained to emulate the behavior of more powerful teacher models. Despite differences in latent space and architectural design, the internal representations of the student and teacher exhibit structurally similar semantic patterns (Fig.~\ref{fig:feature_vis}). This suggests that, despite variations in feature encoding, both models capture and organize semantic information in a comparable manner. Leveraging this structural alignment, we can effectively distill knowledge from the teacher to the student, enabling the latter to inherit the teacher’s superior semantic understanding and generative capabilities, even within a different latent space.

Video frames $\mathbf{v} \in \mathbb{R}^{N \times H \times W \times 3}$ are first encoded by the VAE of both the teacher and student models, producing latent representations $\mathbf{z_{tea}} \in \mathbb{R}^{n_1 \times h_1 \times w_1 \times c_1}$ and $\mathbf{z_{stu}} \in \mathbb{R}^{n_2 \times h_2 \times w_2 \times c_2}$, respectively.
Text embeddings, extracted from their respective text encoders, are injected into the models; we omit them from the notation below for simplicity.
The noisy input latent to teacher $\mathbf{z_{tea}^t}$ is obtained by applying Gaussian noise to $\mathbf{z_{tea}}$ following:
\begin{equation} \label{eq:add_noise} \mathbf{z_\text{tea}^t} = (1 - \sigma(t)) \cdot \mathbf{z_\text{tea}} + \sigma(t) \cdot \mathbf{\epsilon}, \quad \mathbf{\epsilon} \sim \mathcal{N}(0, I), 
\vspace{-5pt}\end{equation}
where $\sigma(t)$ controls the noise level at diffusion timestep $t$, which is randomly sampled following flow matching-based models~\cite{yu2024scaling, hacohen2024ltx}.
The noisy teacher latent $\mathbf{z_{tea}^t}$ first passes through a patch embedding layer, followed by all DiT blocks in the teacher model, ultimately producing the teacher guidance feature $\mathbf{f_{tea}} \in \mathbb{R}^{n_1 \times \frac{h_1}{2} \times \frac{w_1}{2} \times c_1'}$ of the penultimate layer. 
Similarly, the student latent is noised using the same process as in Eq.~\ref{eq:add_noise}, then passed through the student DiT diffusion network, yielding the student feature representation $\mathbf{f_\text{stu}} \in \mathbb{R}^{n_2 \times h_2 \times w_2 \times c_2'}$.
To align the feature representations of the student model with those of the teacher's, we define the distillation loss as
\begin{equation} 
\mathcal{L}_{\text{dis}}(\theta, \phi) := -\mathbb{E}_{\mathbf{v}, \epsilon, t} \left[  \text{sim}(\mathbf{f_\text{tea}}, p_\phi(\mathbf{f_\text{stu}})) \right], 
\end{equation}
where $p_\phi$ is a projector network consisting of spatial and temporal convolutions with interpolation  layers to align the number of tokens in the student feature space with that of the teacher, and maps the student’s feature space to the teacher's.
In addition to the distillation loss $\mathcal{L}_{\text{dis}}$, the student model is also optimized with a diffusion loss as
\begin{equation} 
\label{eq:diff_loss}
\mathcal{L}_{\text{diff}}(\theta) = \mathbb{E}_{\mathbf{v}, \epsilon, t} \left[ \left| \mathbf{\epsilon} - \hat{\mathbf{\epsilon}}_\theta(\mathbf{z_{\text{stu}}^t}, t) \right|_2^2 \right], 
\end{equation}
The final objective is formulated as
\begin{equation} 
\mathcal{L}(\theta, \phi) = \lambda_{\text{dis}} \mathcal{L}_{\text{dis}}(\theta, \phi) + \mathcal{L}_{\text{diff}}(\theta).
\end{equation}

\noindent\textbf{Discussion.} We investigate various strategies for selecting the diffusion timestep $t$ for the teacher and student models. An initial approach involves randomly sampling $t$ for the student while fixing the teacher’s timestep at the final denoising step. This design is motivated by the visualization results in Fig.~\ref{fig:feature_vis}, which show that the teacher’s features at the final step exhibit the most well-defined structural patterns. However, this strategy yields limited performance gains, likely because the teacher’s final-step representations are already close to the clean data distribution. As a result, the distillation process becomes similar to direct data fine-tuning, offering limited insight into the intermediate denoising dynamics. In contrast, synchronizing the teacher and student at the same timestep $t$ leads to notable improvements in generation quality. This suggests that aligning their features across corresponding diffusion steps provides richer and more fine-grained supervision, enabling the student to learn not only the clean end-state representation but also the teacher's progressive denoising trajectory.

\label{sec:method_distill}
\begin{table*}[t]
\centering
\scriptsize
% \tiny
\setlength{\tabcolsep}{2pt}
\renewcommand{\arraystretch}{1.0}
\resizebox{1\textwidth}{!}{
\begin{tabular}{l|c|cc|cccccccccccccccc}
   \textbf{Method}  & 
\makecell[bc]{\rotatebox{90}{\begin{tabular}[c]{@{}l@{}}Total\\ Score\end{tabular}}} & 
\makecell[bc]{\rotatebox{90}{\begin{tabular}[c]{@{}l@{}}Quality\\ Score\end{tabular}}} & 
\makecell[bc]{\rotatebox{90}{\begin{tabular}[c]{@{}l@{}}Semantic\\ Score\end{tabular}}} & 
\makecell[bc]{\rotatebox{90}{\begin{tabular}[c]{@{}l@{}}Subject\\ Consistency\end{tabular}}} & 
\makecell[bc]{\rotatebox{90}{\begin{tabular}[c]{@{}l@{}}Background\\ Consistency\end{tabular}}} & 
\makecell[bc]{\rotatebox{90}{\begin{tabular}[c]{@{}l@{}}Temporal\\ Flickering\end{tabular}}} & 
\makecell[bc]{\rotatebox{90}{\begin{tabular}[c]{@{}l@{}}Motion\\ Smoothness\end{tabular}}} & 
\makecell[bc]{\rotatebox{90}{\begin{tabular}[c]{@{}l@{}}Dynamic\\ Degree\end{tabular}}} & 
\makecell[bc]{\rotatebox{90}{\begin{tabular}[c]{@{}l@{}}Aesthetic\\ Quality\end{tabular}}} & 
\makecell[bc]{\rotatebox{90}{\begin{tabular}[c]{@{}l@{}}Imaging\\ Quality\end{tabular}}} & 
\makecell[bc]{\rotatebox{90}{\begin{tabular}[c]{@{}l@{}}Object\\ Class\end{tabular}}} & 
\makecell[bc]{\rotatebox{90}{\begin{tabular}[c]{@{}l@{}}Multiple\\ Objects\end{tabular}}} & 
\makecell[bc]{\rotatebox{90}{\begin{tabular}[c]{@{}l@{}}Human\\ Action\end{tabular}}} & 
\makecell[bc]{\rotatebox{90}{\begin{tabular}[c]{@{}l@{}}Color\end{tabular}}} & 
\makecell[bc]{\rotatebox{90}{\begin{tabular}[c]{@{}l@{}}Spatial\\ Relationship\end{tabular}}} & 
\makecell[bc]{\rotatebox{90}{\begin{tabular}[c]{@{}l@{}}Scene\end{tabular}}} & 
\makecell[bc]{\rotatebox{90}{\begin{tabular}[c]{@{}l@{}}Appearance\\ Style\end{tabular}}} & 
\makecell[bc]{\rotatebox{90}{\begin{tabular}[c]{@{}l@{}}Temporal\\ Style\end{tabular}}} & 
\makecell[bc]{\rotatebox{90}{\begin{tabular}[c]{@{}l@{}}Overall\\ Consistency\end{tabular}}} \\
    \toprule

HunyuanVideo & 83.24 &85.09 &75.82&97.37&97.76&99.44&98.99&70.83&60.86&67.56&86.10&68.55&94.40&91.60&68.68&53.88&19.80&23.89&26.44 \\
Vchitect(VEnhancer) & 82.24 & 83.54 & 77.06 & 96.83 & 96.66 & 98.57 & 98.98 & 63.89 & 60.41 & 65.35 & 86.61 & 68.84 & 97.20 & 87.04 & 57.55 & 56.57 & 23.73 & 25.01 & 27.57 \\
CogVideoX-1.5 & 82.17 & 82.78 & 79.76 & 96.87 & 97.35 & 98.88 & 98.31 & 50.93 & 62.79 & 65.02 & 87.47 & 69.65 & 97.20& 87.55& 80.25& 52.91& 24.89& 25.19& 27.30 \\
CogVideoX-5B & 81.61 & 82.75 & 77.04 & 96.23 & 96.52& 98.66& 96.92& 70.97 & 61.98 & 62.90& 85.23& 62.11& 99.40& 82.81& 66.35& 53.20& 24.91 & 25.38 & 27.59 \\
CogVideoX-2B & 81.57 & 82.51 & 77.79& 96.42 & 96.53 & 98.45 & 97.76 & 58.33 & 61.47 & 65.60 & 87.81 & 69.35 & 97.00 & 86.87& 54.64& 57.51& 24.93 & 25.56 & 28.01 \\
Mochi-1 & 80.13 &82.64 &70.08 & 96.99 & 97.28 & 99.40 & 99.02 & 61.85 & 56.94 & 60.64 & 86.51 & 50.47 & 94.60 & 79.73 & 69.24 & 36.99 & 20.33& 23.65 & 25.15 \\
LTX-Video & 80.00 & 82.30 & 70.79 & 96.56 & 97.20 & 99.34 & 98.96 & 54.35 & 59.81 & 60.28 & 83.45 & 45.43 & 92.80 & 81.45 & 65.43 & 51.07 & 21.47 & 22.62 & 25.19 \\
OpenSora-1.2 & 80.00 & 82.30 & 70.79 & 96.56 & 97.20 & 99.34 & 98.96& 54.35 & 59.81& 60.28 & 83.45 & 45.43 & 92.80 & 81.45 & 65.43 & 51.07 & 21.47 &22.62 & 25.19 \\
OpenSoraPlan-V1.1 &  78.00 & 80.91 & 66.38 & 95.73 & 96.73 & 99.03 & 98.28 & 47.72 & 56.85 & 62.28 & 76.30 & 40.35 & 86.80 & 89.19 & 53.11 & 27.17 & 22.90 & 23.87 & 26.52 \\
\midrule
\rowcolor{gray!10}\textbf{Turbo2K(Ours)} 
&82.78&84.91&74.24&96.77&97.20&99.20&98.86&74.65&61.78&65.62&85。82&53.58&95.20&86.95&75.44&51.08&21.01&22.41&26.93\\

\bottomrule 
\end{tabular}
}
\caption{Comparison with state-of-the-art open-source models on VBench-Long benchmark.}
\vspace{-0.1in}
\label{tab:vbench}
\end{table*}

\begin{table}[htbp]
\centering
\resizebox{1\linewidth}{!}{
\renewcommand{\arraystretch}{1.2}
% \begin{adjustbox}{max width=0.8\textwidth}
\begin{tabular}{lccc}
\toprule
\multirow{2}{*}{\textbf{Model}} & \multirow{2}{*}{\textbf{Params}} & \multicolumn{2}{c}{\textbf{Inference Time (s/step)}} \\
\cmidrule(lr){3-4}
 & & \textbf{49 frames@2K} & \textbf{121 frames@2K} \\
\midrule
CogvideoX & 2B & 94.68 & OOM \\
CogvideoX & 5B & 220.12 & OOM \\
CogvideoX 1.5 & 5B & 87.82 & OOM \\
Mochi & 10B & 105.09 & OOM \\ 
Hunyuan Video & 13B & 132.70 & OOM \\ \midrule
\rowcolor{gray!10}\textbf{Torbo2k (Ours)} & \textbf{4B} & \textbf{5.08} & \textbf{22.80} \\
\bottomrule
\end{tabular}
% \end{adjustbox}
}
\caption{Comparison of model parameters and inference Time at 2K Resolution. OOM denotes out of memory.}
\label{tab:time_size}
\end{table}

\subsection{Two-stage Synthesis}
\label{sec:method_twostage}
While knowledge distillation allows the student model to inherit the teacher’s semantic understanding and denoising trajectory, directly synthesizing 2K-resolution videos remains a significant challenge due to the increased complexity of texture generation and semantic planning. To address this, we introduce a two-stage synthesis framework in which high-resolution video generation is guided by a low-resolution synthesis process. A naive solution is to first generate a low-resolution video and subsequently condition a high-resolution generation on it, akin to generative super-resolution techniques~\cite{he2024venhancer, zhou2024upscale, wang2024exploiting}. However, such approaches require explicit decoding of the LR output and re-encoding it as a condition during inference, leading to considerable computational overhead. This issue is exacerbated when processing high-resolution inputs, which typically demand tiled encoding to meet memory constraints. Moreover, this decoupled paradigm tends to induce excessive reliance on the LR input, constraining the model’s capacity to enrich fine details. Consequently, the HR output often exhibits minimal enhancement over its LR counterpart.

To overcome these limitations, we propose a feature-based guidance approach rather than direct pixel-space conditioning. Specifically, we extract intermediate feature representations from multiple DiT blocks during LR video synthesis, and use them as semantic guidance for HR generation. This strategy eliminates the need for redundant LR decoding and encoding steps during inference while providing rich structural information to the HR model.

As illustrated in the right part of Fig.~\ref{fig:method}, we extract multi-level feature representations from selected DiT blocks, denoted as $\mathcal{G} = \{ \mathbf{g}_i \mid i \in \mathcal{D} \}$, 
where $\mathcal{D}$ represents the set of selected DiT block indices. 
The timestep for LR guidance extraction is fixed at the final diffusion step, as the feature representation at this stage is more stable and semantically clear, shown in Fig.~\ref{fig:feature_vis}, providing effective structural information for HR generation.
During HR video synthesis, the extracted guidance features are injected into their corresponding DiT blocks in the HR generation branch via fusion blocks, as shown in Fig.~\ref{fig:method}. 
Since the LR features $\mathbf{g}$ have a lower spatial resolution than the HR features, we first upsample them using interpolation, followed by local convolutional operations to match their resolution to that of the HR features. 
The upsampled guidance features $\mathbf{g'}$ are combined with the HR features $\mathbf{h}$ as
\begin{equation} 
\mathbf{h'}= \text{Norm}(\text{MLP}(\mathbf{h})) + \mathbf{g'},
\end{equation}
The feature $\mathbf{h'}$ is modulated by the time embedding $t$ as
\begin{equation} \mathbf{h''} = \mathbf{h'} \odot (1 + \boldsymbol{\alpha}_t) + \boldsymbol{\beta}_t, \end{equation} where $\boldsymbol{\alpha}_t$ and $\boldsymbol{\beta}_t$ are modulation parameters predicted from the time embedding. The resulting fused feature $\mathbf{h''}$ is then passed to the subsequent DiT blocks.
The LR model remains fixed, while the HR model is initialized with LR parameters and trained with a diffusion loss similar to Eq.~\ref{eq:diff_loss}.
\begin{figure*}[t]
  \centering
   \includegraphics[width=1\linewidth]{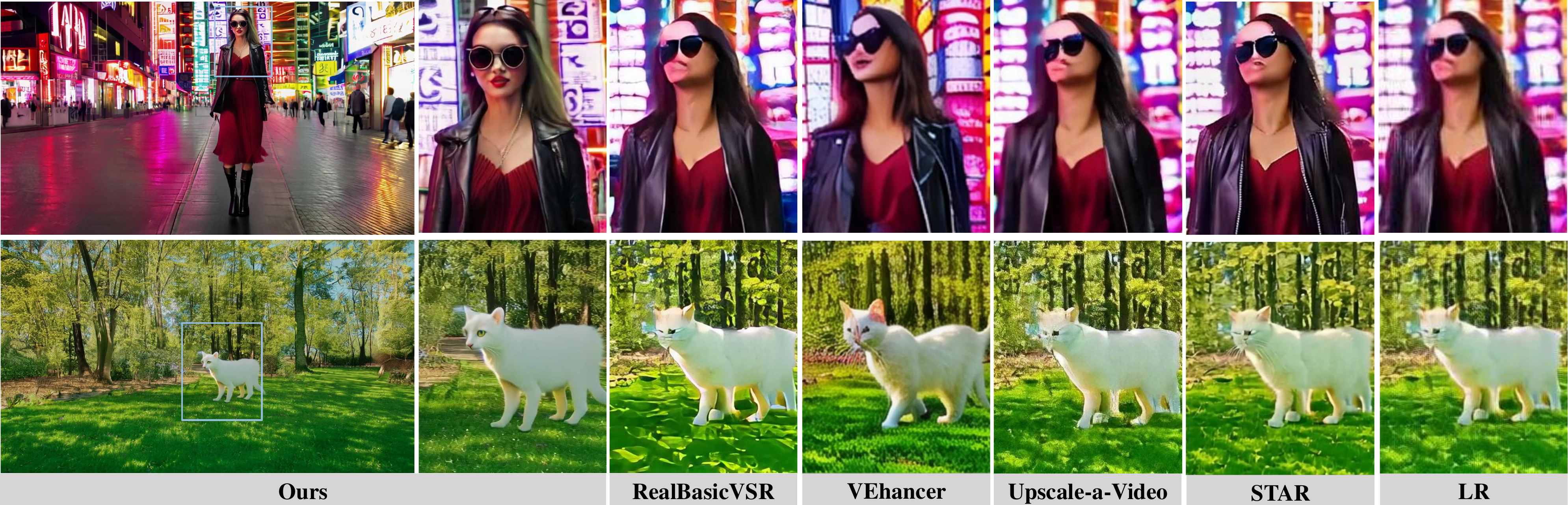}
   \vspace{-0.2in}
   \caption{Visual comparison with video super-resolution methods. Our approach produces high-resolution results with finer details and stronger semantic coherence. Unlike existing VSR methods that heavily depend on the LR input, our method maintains semantic alignment while refining structural details, ensuring improved fidelity and consistency in HR synthesis.}
   \vspace{-0.1in}
   \label{fig:vsr_compare}
\end{figure*}
\vspace{-0.2in}

\section{Experiment}
%\vspace{-0.1in}
We use HunyuanVideo (13B) as the teacher model for its strong generative capability and LTX-Video (2B) as the student for its high compression ratio and efficiency. 
Please refer to the supplementary file for further implementation details.
In the next, Sec.~\ref{sec:compare_study} compares Turbo2K with T2V and VSR models, while Sec.~\ref{sec:abla_study} analyzes distillation and two-stage synthesis.

\subsection{Comparative Study}
\label{sec:compare_study}
%\vspace{-0.1in}
\noindent\textbf{Comparison with T2V models}.
We evaluate our model’s generative performance on VBench-Long~\cite{huang2024vbench}, a benchmark designed to measure both visual quality and semantic coherence in video generation. As shown in Table~\ref{tab:vbench}, our Turbo2K model (4B parameters) achieves high visual fidelity and aesthetic quality. Qualitative examples in Fig.~\ref{fig:teaser} and the supplementary materials further demonstrate the semantic consistency and visual realism of the generated videos. Remarkably, Turbo2K surpasses CogVideoX-5B in performance while being more lightweight and significantly faster, underscoring the efficiency of our method.

Table~\ref{tab:time_size} compares the time complexity and model size across different methods. To evaluate computational efficiency, we report the inference time per denoising step on an A100 GPU. For 2K-resolution video generation with 49 frames, Turbo2K achieves a runtime of just 5.08 seconds per step—nearly 20× faster than CogVideoX-2B and over 40× faster than CogVideoX-5B. Notably, existing methods are unable to generate 121-frame 2K-resolution videos within the memory constraints of a single A100 GPU. In contrast, Turbo2K successfully completes this task with an average runtime of just 22.8 seconds per denoising step, demonstrating superior scalability and efficiency in high-resolution, long-form video generation.

\begin{figure*}[t]
  \centering
   \includegraphics[width=1\linewidth]{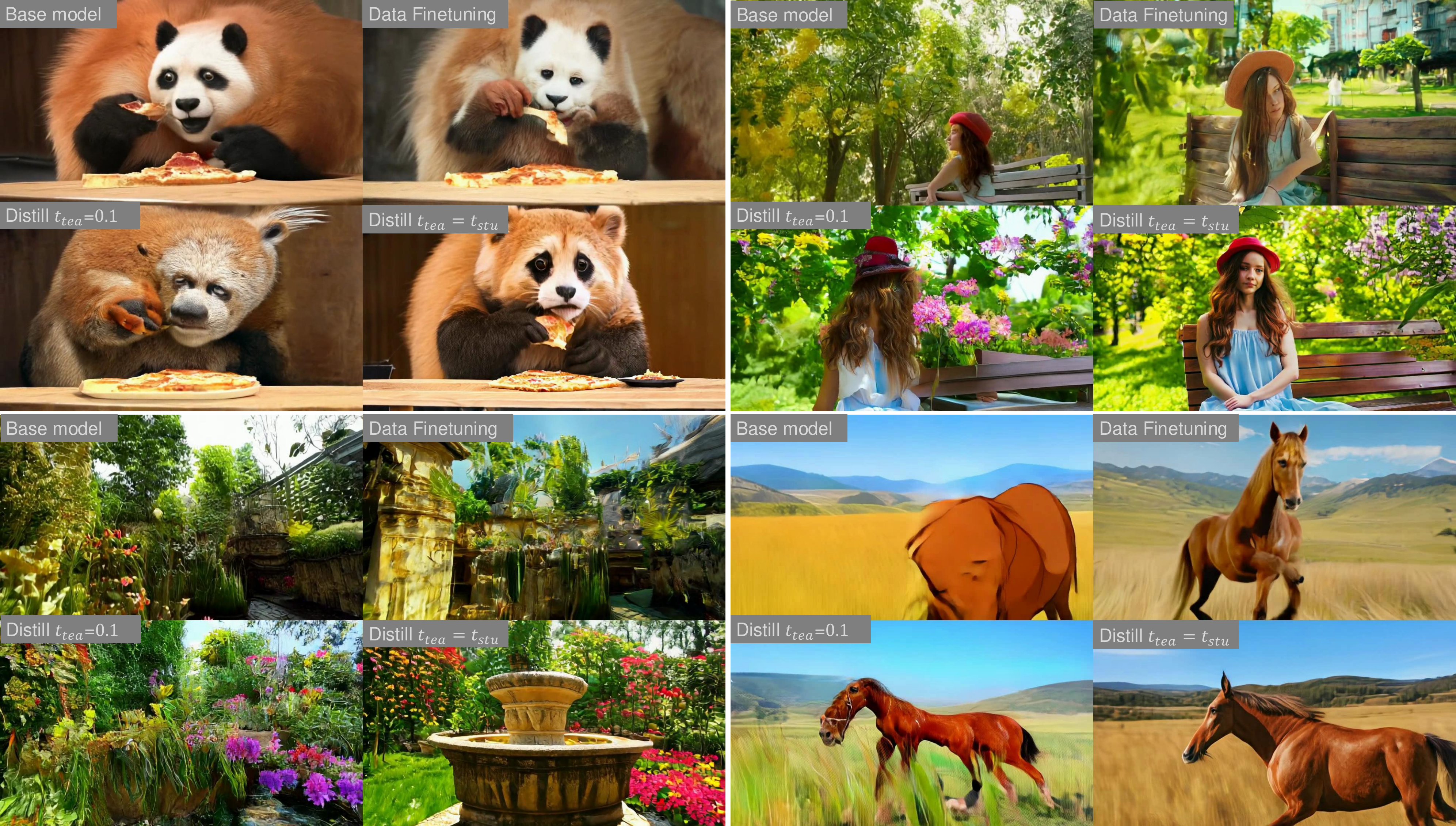}
   \caption{Visual comparison between data fine-tuning and the proposed distillation method. Please refer to our supplementary file for comparison in video format.}
   \vspace{-0.1in}
   \label{fig:distill_compare}
\end{figure*}

\begin{figure*}[t]
  \centering
   \includegraphics[width=1\linewidth]{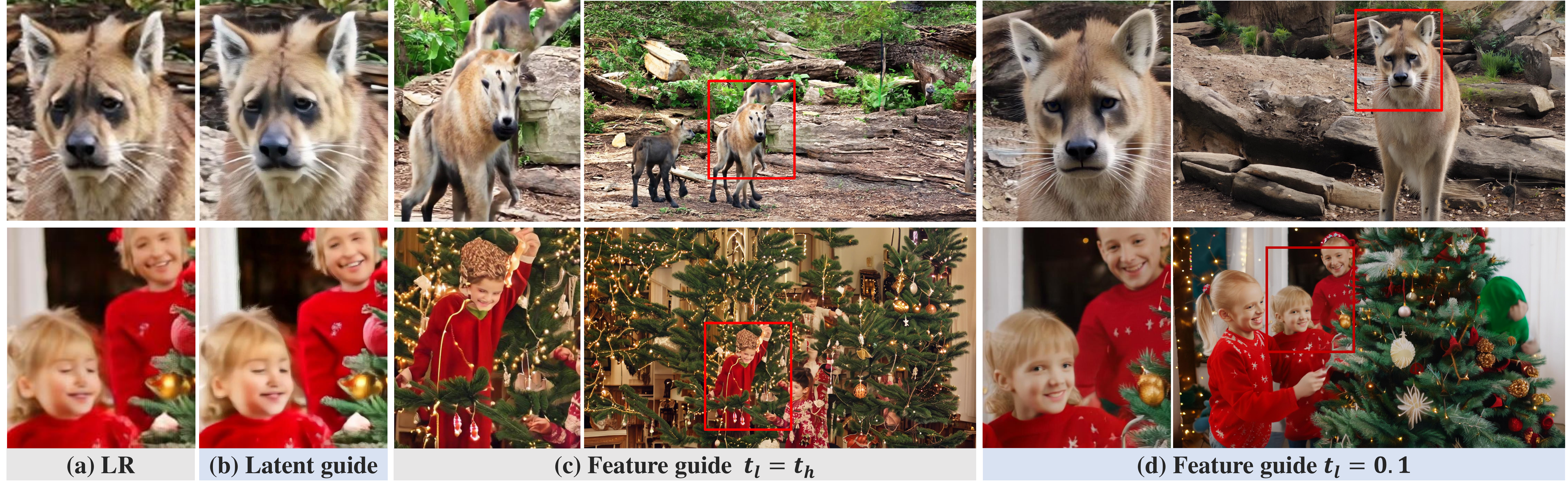}
   \vspace{-0.2in}
   \caption{Visual comparison of different guidance strategies. (a) LR result, (b) Results using latent representation guidance, (c) Feature guidance with timesteps aligned to HR generation, and (d) Feature guidance with $t = 0.1$.}
   \vspace{-0.05in}
   \label{fig:abla_up2k}
\end{figure*}

\noindent\textbf{Comparison with VSR models}. 
We thoroughly compare our method with representative video super-resolution (VSR) approaches, including RealBasicVSR~\cite{chan2022investigating}, Upscale-a-Video~\cite{zhou2024upscale}, VEhancer~\cite{he2024venhancer}, and STAR~\cite{xie2025star}. 
We use MUSIQ~\cite{ke2021musiq}, NIQE~\cite{mittal2012making}, and CLIPIQA~\cite{wang2023exploring} to evaluate frame quality and detail richness; the technical and aesthetic score of DOVER~\cite{wu2023exploring} to assess video visual quality and consistency; and the consistency metric from~\cite{huang2024vbench} to measure semantic alignment.
The quantitative results are presented in Table~\ref{tab:vsr_compare}, while qualitative comparisons are illustrated in Fig.~\ref{fig:vsr_compare}.
RealBasicVSR exhibits limited ability to generate details beyond those present in the low-resolution input. Diffusion-based approaches—such as Upscale-a-Video, VEhancer, and STAR—can enhance LR content by introducing additional details. However, their outputs often remain overly dependent on the initial LR generation, leading to unresolved structural inconsistencies, such as distortions in facial regions (\textit{e.g.}, a woman's mouth or a cat’s face). In contrast, our method delivers detail-rich and semantically consistent results while preserving high computational efficiency.
%
%\vspace{-5pt}
\subsection{Ablation Study}
%\vspace{-0.1in}
\label{sec:abla_study}
To efficiently evaluate various design choices, we select 120 text prompts containing detail-rich descriptions to facilitate a rapid and reliable assessment of model performance.

\noindent\textbf{Heterogeneous model distillation.}
We analyze the impact of different model fine-tuning configurations on video generation quality. 
We first evaluate data fine-tuning, where the student model is directly trained using the flow matching loss, without explicit feature alignment with the teacher model. 
As shown in Table~\ref{tab:abla_distill} and Fig.~\ref{fig:distill_compare}, this approach does not yield significant improvements over the LTX-Video baseline, suggesting that naive fine-tuning is insufficient for enhancing generative performance.

To improve knowledge transfer, we introduce a distilled base model, where the student network’s feature representations are aligned with those of the teacher model. We experiment with two configurations: \textbf{(1)} Fixed Teacher Timestep ($t=0.1$): The student model is trained to match the teacher model’s features extracted at $t=0.1$. \textbf{(2)} Synchronized Timesteps: The teacher model’s timestep is dynamically synchronized with that of the student model during training. 
As shown in Table~\ref{tab:abla_distill} and Fig.~\ref{fig:distill_compare}, the fixed-step $t_{tea}=0.1$ setting yields limited gains, as the teacher’s final-step features closely resemble clean data—similar to data fine-tuning. In contrast, synchronizing timesteps ($t_{tea} = t_{stu}$) substantially improves generation quality, demonstrating the effectiveness of stepwise feature alignment.

\begin{table}[!t]
\centering
\begin{adjustbox}{width=\hsize}
\setlength{\tabcolsep}{3pt}
\renewcommand{\arraystretch}{1.2} % Adjust row height
\normalsize
\begin{tabular}{lcccccc}
\toprule
& \multicolumn{3}{c}{\textbf{Frame Quality}} & \multicolumn{3}{c}{\textbf{Video Quality}} \\
\cmidrule(lr){2-4} \cmidrule(lr){5-7} 
\textbf{Method} & \textbf{MUSIQ$\uparrow$} & \textbf{NIQE$\downarrow$} & \textbf{CLIPIQA$\uparrow$} & \textbf{Tech.$\uparrow$} & \textbf{Aesth.$\uparrow$} & \textbf{Seman.$\uparrow$} \\
\midrule
RealBasicVSR &\underline{51.48} &\underline{4.64}&\underline{0.44} &99.41& 10.81&\underline{24.41} \\
Upscale-a-Video&43.56 &5.34 &0.42 &99.39&12.15&22.35\\
VEhancer&34.79 &6.39 &0.33&\textbf{99.42}&7.66&24.09\\
STAR &41.20 &4.69 &0.37&99.40&\textit{12.58}&23.13\\
\rowcolor{gray!10}\textbf{Ours}&\textbf{52.45} &\textbf{4.43} &\textbf{0.47}&\textbf{99.42}&\textbf{13.70}&\textbf{24.77}\\
\bottomrule
\end{tabular}
\end{adjustbox}
\vspace{-0.1in}
 \caption{ Frame and video quality comparison between state-of-the-art VSR methods and our Turbo2K. }
\vspace{-0.1in}
\label{tab:vsr_compare}
\end{table}

\begin{table}[!t]
\centering
\setlength{\tabcolsep}{3pt}
\renewcommand{\arraystretch}{1.2} % Adjust row height
\Large%\scriptsize
\resizebox{1\linewidth}{!}{
\begin{tabular}{lcccccc}
\toprule
& \multicolumn{3}{c}{\textbf{Frame Quality}} & \multicolumn{3}{c}{\textbf{Video Quality}} \\
\cmidrule(lr){2-4} \cmidrule(lr){5-7} 
\textbf{Method} & \textbf{MUSIQ$\uparrow$} & \textbf{NIQE$\downarrow$} & \textbf{CLIPIQA$\uparrow$} & \textbf{Tech.$\uparrow$} & \textbf{Aesth.$\uparrow$} & \textbf{Seman.$\uparrow$} \\
\midrule
Basic LTX & 58.80 & 4.54 & \underline{0.42} & 10.72 & \underline{99.28} & 22.47  \\
Data fine-tuning & 60.36 & \underline{4.43} & 0.40 & 10.64 & 99.02 & 22.59  \\
Distill $t_\text{tea}$ = 0.1 & \underline{61.12} & 4.53 & 0.41 & \underline{10.87} & 99.15 & \underline{22.62}  \\
\rowcolor{gray!10}Distill $t_\text{tea}=t_{stu}$  & \textbf{64.73} & \textbf{4.15} & \textbf{0.49} & \textbf{11.54} & \textbf{99.46} & \textbf{24.48} \\
\bottomrule
\end{tabular}
}
\vspace{-0.1in}
\caption{Frame and video quality comparison between baseline LTX and various fine-tuning methods. The best results are marked in \textbf{bold} and the second-best in \underline{underline}.}
\vspace{-0.1in}
\label{tab:abla_distill}
\end{table}

\begin{table}[!t]
\centering
\setlength{\tabcolsep}{3pt}
\renewcommand{\arraystretch}{1.2} % Adjust row height
%\scriptsize
\resizebox{1\linewidth}{!}{
\begin{tabular}{lcccccc}
\toprule
& \multicolumn{3}{c}{\textbf{Frame Quality}} & \multicolumn{3}{c}{\textbf{Video Quality}} \\
\cmidrule(lr){2-4} \cmidrule(lr){5-7} 
\textbf{Guide} & \textbf{MUSIQ$\uparrow$} & \textbf{NIQE$\downarrow$} & \textbf{CLIPIQA$\uparrow$} & \textbf{Tech.$\uparrow$} & \textbf{Aesth.$\uparrow$} & \textbf{Seman.$\uparrow$} \\
\midrule
\large Latent & \large49.42 & \large 5.04 & \large 0.391 & \large 9.72 & \large 98.40 & \large 24.04 \\
\large Feature $t=t_{hr}$ & \large 49.16 & \large 4.77 & \large 0.405 & \large 9.13 & \large 99.04 & \large 18.53 \\
\rowcolor{gray!10}\large Feature $t=0.1$ & \large \textbf{52.98} & \large \textbf{4.03} & \large \textbf{0.422} & \large \textbf{12.20} & \large \textbf{99.45} & \large \textbf{24.60} \\
\bottomrule
\end{tabular}
}
\vspace{-0.1in}
\caption{Frame and video quality comparison among various methods of extracting LR guidance. The best results are highlighted in \textbf{bold}.}
\vspace{-0.2in}
\label{table:2k_up}
\end{table}
\noindent\textbf{Two-stage synthesis.}
We compare two strategies for guiding the HR generation: (1) using the LR latent representation and (2) using multi-level semantic features. The first approach, follwing~\cite{chan2022investigating, he2024venhancer, xie2025star}, applies hand-crafted degradation to HR videos to obtain LR inputs, from which latent representations are extracted to guide the HR model. As shown in Fig.~\ref{fig:abla_up2k}, this latent-guided strategy tends to make the HR model overly reliant on the LR input, limiting its ability to generate richer, fine-grained details. Additionally, it introduces extra computational overhead due to the need for separate LR decoding and encoding.

For the feature-based guidance approach, we explore two settings: (1) synchronizing the timesteps of LR feature extraction with those of HR generation ($t=t_{hr}$), and (2) using late-step LR features extracted at $t=0.1$. As shown in Fig.~\ref{fig:abla_up2k} and Table~\ref{table:2k_up}, the synchronized setting results in lower semantic coherence, as the early-stage LR features lack stable structural patterns (see Fig.~\ref{fig:feature_vis}) and thus provide unreliable guidance. Conversely, extracting LR features at a late diffusion step offers semantically rich and structurally refined signals that effectively guide HR synthesis. This leads to enhanced detail generation and improved semantic consistency in the final output.

\section{Limitation and Future Work}
\label{supp:sec_limitation}
While Turbo2K demonstrates strong efficiency and quality in 2K video generation, its performance remains challenged in highly dynamic or visually complex scenes. In particular, the model occasionally produces unnatural or inconsistent hand poses, reflecting difficulties in synthesizing fine-grained motion details. These limitations are likely attributable to the constrained model capacity and limited diversity in high-quality training data. Future work will investigate scaling the model and expanding the dataset to enhance its generalization and fidelity in such challenging scenarios.

Moreover, the current VAE architecture does not support temporally partitioned decoding, which poses a bottleneck for generating long-duration videos due to substantial memory demands. Addressing this constraint through a more efficient, sequential VAE decoding scheme will be essential for enabling scalable and temporally consistent high-resolution video synthesis.

%\vspace{-0.1in}
\section{Conclusion}
%\vspace{-0.1in}
We present Turbo2K, an ultra-efficient framework that combines heterogeneous model distillation and a two-stage synthesis to achieve high-quality 2K video generation. Our distillation approach effectively transfers generative capabilities from large-scale teacher models, while two-stage synthesis ensures fine-grained details and structural coherence.
Ablation studies confirm that synchronizing teacher-student timesteps enhances knowledge transfer, and feature-based LR guidance improves HR synthesis by preserving details and consistency. Evaluations on VBench show that Turbo2K outperforms larger models like CogVideoX-5B, generating 2K videos up to 40× faster with superior quality.
%Turbo2K leverages distillation and feature-guided synthesis to achieve strong performance and scalability in high-quality video generation.

{\small
\bibliographystyle{ieee_fullname}
\bibliography{egbib}
}
\clearpage

\clearpage
\section{Supplementary}
\noindent\textbf{Overview.} 
This supplementary document provides additional implementation details in Sec.~\ref{supp:sec_implement}, further analysis of LR guidance strategies and additional visual results in Sec.~\ref{supp:sec_addition}. 

\subsection{Implementation Details}
\label{supp:sec_implement}
To train our model, we curated a high-resolution video dataset with full copyright ownership. The dataset comprises approximately 410K high-quality videos covering a diverse range of scenes and categories, with the majority of samples available in 4K resolution. Each video is labeled using ShareGPT4Video~\cite{chen2024sharegpt4video}, providing rich textual descriptions to facilitate text-to-video training.
To enhance the diversity of training samples, we adopt a mixed training approach that combines videos and images at a 2:1 ratio. This strategy ensures that the model effectively learns both temporal dynamics from videos and high-quality spatial details from images, contributing to improved generative performance. 
All experiments are conducted using the Adam optimizer with a learning rate of $10^{-4}$.

\noindent\textbf{Progressive training strategy.}
To optimize training efficiency and stabilize convergence, we employ a progressive training strategy where the model is trained at incrementally increasing resolutions. During the heterogeneous model distillation stage, the student model is first trained at a resolution of $544\times 960$, allowing it to effectively inherit knowledge from the teacher model while maintaining computational efficiency.
For the two-stage synthesis, the HR model is initially trained at a resolution of $49 \times 1440 \times 2560$ for 5K iterations, enabling it to establish a coarse high-resolution structure. Subsequently, the model is further fine-tuned at a resolution of $121 \times 1440 \times 2560$ for an additional 8K iterations, allowing for enhanced detail refinement and temporal consistency.
Both LR and HR models consist of 28 DiT blocks, with LR guidance extracted at block indices ${0, 7, 14, 21}$. Each extracted feature is fused with its corresponding HR feature through a fusion block.

\subsection{Addition Results}
\label{supp:sec_addition}
This section first provides further explanations on the timestep configurations for extracting LR guidance in Sec.~\ref{supp:sec_additionvisual}, followed by additional generated results of Turbo2K in Sec.~\ref{supp:sec_more_visual}. We also recommend watching the videos provided in the supplementary file for a more comprehensive evaluation.

Additionally, we present a visual comparison in video format for heterogeneous model distillation, including results from the LTX baseline, pure data fine-tuning, distillation with a fixed teacher timestep at the final step, and distillation where the teacher timestep is aligned with the student model. The comparison demonstrates that pure data fine-tuning yields limited improvements, while fixing the teacher’s timestep at the final step provides insufficient supervision, as the teacher’s features at this stage closely resemble clean data. In contrast, aligning teacher and student timesteps during distillation better preserves the denoising trajectory, leading to superior generative quality and improved semantic coherence.

\begin{figure}[t]
  \centering
   \includegraphics[width=1\linewidth]{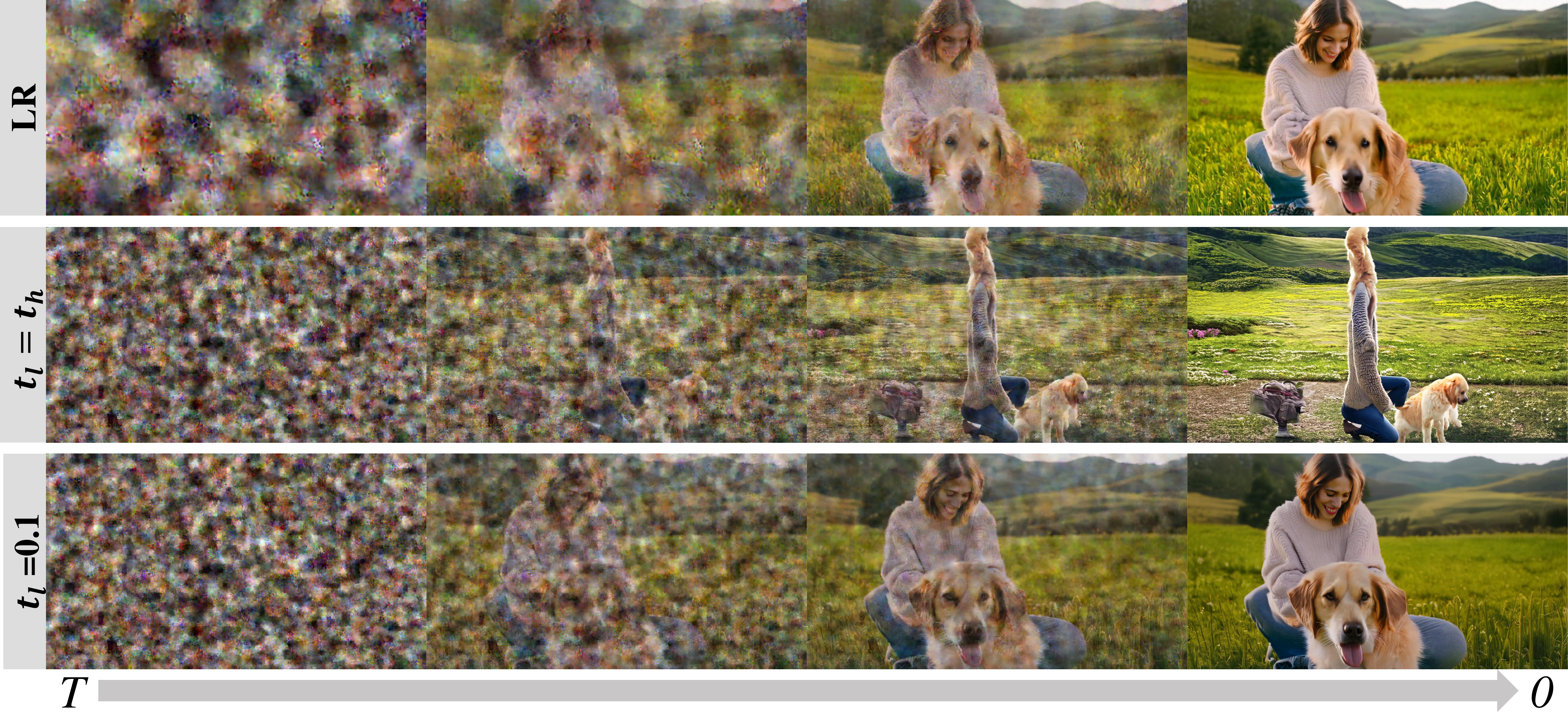}
   \caption{Comparison of LR results and HR results guided by LR-based feature guidance across timesteps. }
   \label{fig:lrtimestep_compare}
\end{figure}
\subsubsection{Analysis of LR Guidance Strategies }
To further analyze the effect of LR guidance on HR generation, we provide visual comparisons of intermediate frames across different guidance configurations in Fig.~\ref{fig:lrtimestep_compare}. 
We compare standard LR generation, synchronized LR-HR timesteps where LR features are extracted at the same timestep as HR denoising, and final-step LR guidance where LR features are taken from the last denoising step. 
The results indicate that synchronizing LR and HR timesteps leads to unstable HR structures, as early LR features are not yet well-formed, causing HR synthesis to inherit ambiguous details. 
By the time LR features stabilize, HR is already in its final refinement stage, limiting its ability to incorporate structural corrections. 
In contrast, using LR features from the final denoising step provides the most stable and informative guidance, ensuring coherent structural formation in HR synthesis. 
\label{supp:sec_additionvisual}

\subsubsection{More Visual Results of Turbo2K}
\begin{figure*}[t]
  \centering
   \includegraphics[width=1\linewidth]{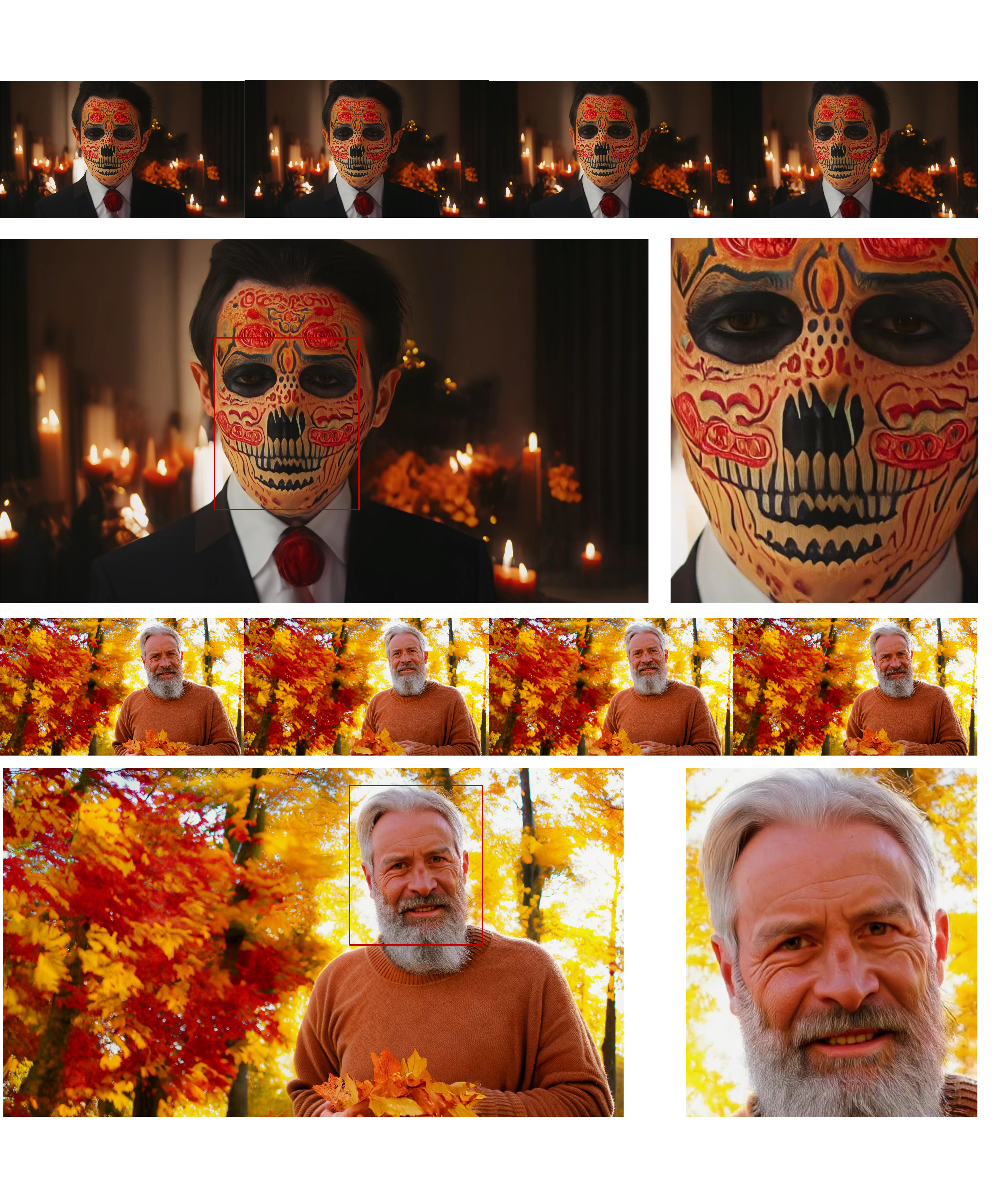}
   \caption{Our Turbo2K generated results}
   \label{supp:fig_man}
\end{figure*}

\begin{figure*}[t]
  \centering
   \includegraphics[width=1\linewidth]{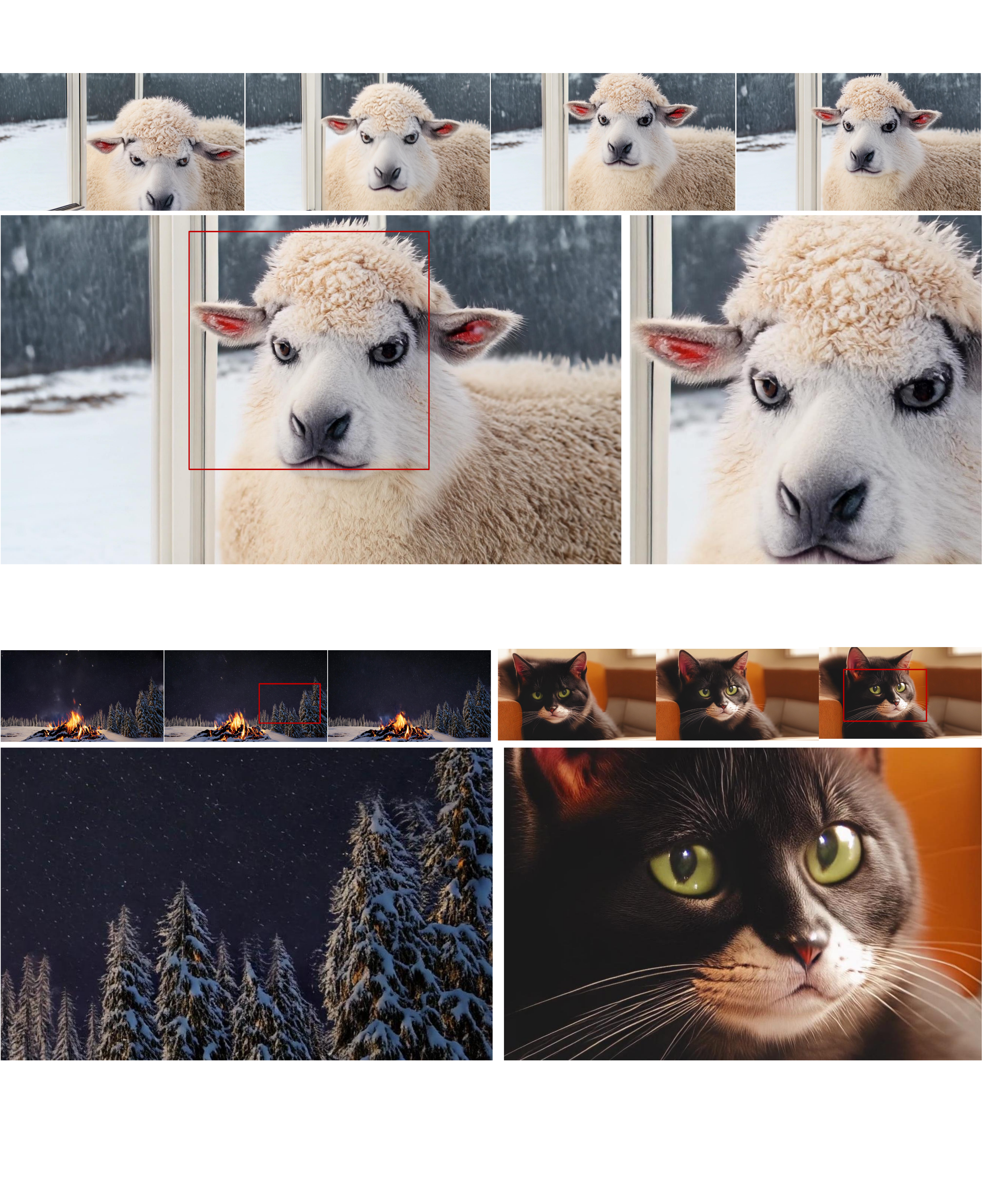}
   \caption{Our Turbo2K generated results}
   \label{supp:animal}
\end{figure*}

\begin{figure*}[t]
  \centering
   \includegraphics[width=1\linewidth]{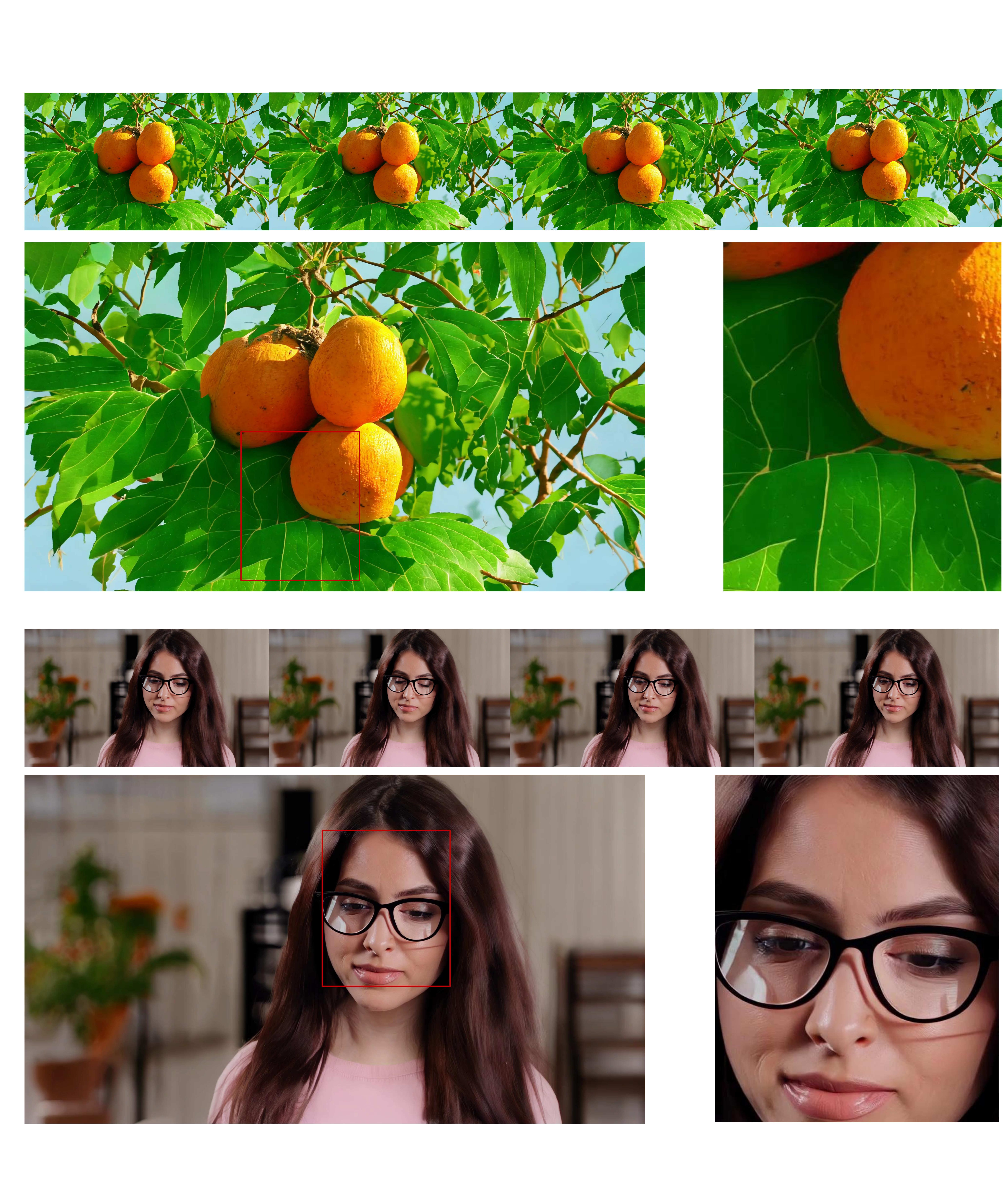}
   \caption{Our Turbo2K generated results}
   \label{supp:woman}
\end{figure*}

We present additional frames generated by Turbo2K in Fig.~\ref{supp:fig_man}, Fig.~\ref{supp:animal}, and Fig.~\ref{supp:woman}, demonstrating rich details, high aesthetic quality, and strong semantic coherence. Additionally, we provide video results in this supplementary file and recommend viewing them for a more comprehensive evaluation.
\label{supp:sec_more_visual}

\end{document}